\documentclass{article}

\usepackage[preprint, nonatbib]{neurips_2022}

\usepackage[utf8]{inputenc} % allow utf-8 input
\usepackage[T1]{fontenc}    % use 8-bit T1 fonts
\usepackage{url}            % simple URL typesetting
\usepackage{booktabs}       % professional-quality tables
\usepackage{amsfonts}       % blackboard math symbols
\usepackage{nicefrac}       % compact symbols for 1/2, etc.
\usepackage{microtype}      % microtypography
\usepackage{xcolor}         % colors
\usepackage{amsmath}
\usepackage{bm}
\usepackage{graphicx}
\usepackage{subfigure}

\def\ve{{\bm{e}}}
\def\vr{{\bm{r}}}
\newcommand{\R}{\mathbb{R}}

\title{AutoWeird:
	Weird Translational Scoring Function
	Identified by Random Search}

\author{%
  Hansi Yang \\
  Department of Computer Science and Engineering \\
  Hong Kong University of Science and Technology \\
  Clear Water Bay, Hong Kong \\
  \texttt{hyangbw@cse.ust.hk} \\
   \And
   Yongqi Zhang \\
   4Paradigm Inc. \\
   Beijing, China \\
   \texttt{zhangyongqi@4paradigm.com} \\
   \And
   Quanming Yao \\
   Department of Electronic Engineering \\
   Tsinghua University \\
   Beijing, China \\
   \texttt{qyaoaa@tsinghua.edu.cn} \\
}

\begin{document}

\maketitle

\begin{abstract}
Scoring function (SF) measures the plausibility of triplets in knowledge graphs. 
Different scoring functions can lead to huge differences in link prediction performances on different knowledge graphs. 
In this report, we describe a weird scoring function 
found by random search on the open graph benchmark (OGB).
This scoring function, called AutoWeird, only uses tail entity and relation in a triplet to compute its plausibility score. 
Experimental results show that AutoWeird achieves top-1 performance on ogbl-wikikg2 data set, 
but has much worse performance than other methods on ogbl-biokg data set. 
By analyzing the tail entity distribution and evaluation protocol of these two data sets, 
we attribute the unexpected success of AutoWeird on ogbl-wikikg2 to inappropriate evaluation and concentrated tail entity distribution. 
Such results may motivate further research on how to accurately evaluate the performance of different link prediction methods for knowledge graphs. 
\end{abstract}

\section{Introduction}

%{\color{blue}
%\begin{itemize}
%\item observation on recent methods of ogbl-wikikg2
%
%\item motivation to use AutoML techniques
%
%\item our method and results:  a weird scoring function is identified
%
%\item analysis of the result:
%\end{itemize}
%}

Scoring function (SF),
which measures the plausibility of triplets 
in knowledge graphs (KGs), 
plays an important role to learn from KGs \cite{wang2017knowledge,ji2021survey}.
To fairly evaluate different SFs,
open graph benchmark (OGB) \cite{hu2020ogb} sets up leaderboards 
to evaluate the different models on large-scale graphs.
Among the two leaderboards for link prediction task in KG,
we observe that the leading models on \texttt{ogbl-wikikg2} 
have similar forms by modeling embeddings as translations in vector spaces.
To analyze the success of these SFs
and try to design a better SF,
we are motivated to use automated machine learning (AutoML) techniques \cite{yao2018automl,zhang2020autosf,zhang2022bilinear}
to search SFs in this type.

Based on the translational models~\cite{bordes2013translating, wang2022interht, long2021triplere, chao2020pairre, zhang2022trans},
we first set up a search space that can cover all these translational models as special cases.
Then,
we conduct random search in this space to explore new designs of scoring functions.
After evaluating $10$ SFs, 
we identified a weird SF which does not contain head entity embedding
but outperforms the existing models.
This is abnormal as triplets in KGs are formed by head entity, relation and tail entity.
By analyzing the distribution of tail entities,
we guess the success of such a weird SF may attribute to the
concentrated distribution and the evaluation protocol
with random negative sampling.

This tech-report is organized as follows.
First,
we introduce the search space,
search algorithm
and the training objective in Section~\ref{sec:method}.
Then in Section~\ref{sec:exp}, 
we show the 
the top-1 performance achieved by
the weird translation-based SF on ogbl-wikikg2
and our analysis on the searched results.

\section{Method}
\label{sec:method}

The scoring function used here is found by automated machine learning techniques~(AutoML)~\cite{hutter2019automated,yao2018automl}. 
We first describe the search space where we find the scoring function (SF) in Section~\ref{ssec:space}.
And our search algorithm is introduced in Section~\ref{ssec:alg}. 

\subsection{Search space}
\label{ssec:space}

Our search space is motivated by translation-based models~\cite{bordes2013translating, wang2022interht, long2021triplere, chao2020pairre, zhang2022trans}
on the ogbl-wikikg2 leaderboard
\footnote{\url{https://ogb.stanford.edu/docs/leader_linkprop/#ogbl-wikikg2}}. 
Our search space is designed as follows:
\begin{itemize}
\item We set up the entity embeddings with two parts $\ve_0, \ve_1$, 
relation embeddings with three parts $\vr_0, \vr_1, \vr_2$. 
All these vectors have same dimension $d$, i.e. $\ve_0, \ve_1, \vr_0, \vr_1, \vr_2 \in \R^{d}$.

\item The scoring function is given by combining the the different components,
i.e., $s(h, r, t) = - \| f(\ve_0^h, \ve_1^h, \vr_0, \vr_1, \vr_2, \ve_0^t, \ve_1^t) \|_1$, 
where $\ve_0^h, \ve_1^h$ denote the embeddings for head entity $h$,
$\ve_0^t, \ve_1^t$ denote the embeddings for tail entity $t$, 
$\vr_0, \vr_1, \vr_2$ denote the embeddings for relation $r$,
and $\| \cdot \|_1$ is the $\ell_1$ norm. 

\item 
The function $f(\cdot)$ outputs a vector in $\R^{d}$, and is computed by adding/subtracting first-order terms ($\ve_0^h, \ve_1^h, \vr_0, \vr_1, \vr_2, \ve_0^t, \ve_1^t$) and/or second-order terms (element-wise product between any of these two vectors), e.g., $\bm e_0^h\circ \bm e_1^t$ 
and
$\bm e_0^h\circ \bm r_1$,
for entity and relation embeddings. 
\end{itemize}

We can see that many translation-based SFs that are already on the OGB leaderboard~\cite{hu2020ogb} can all be covered in this search space. 
Some examples are listed in Table~\ref{tab:related}. 

\begin{table*}[!ht]
\small
\caption{Example knowledge graph embedding models that can be covered by our search space.}
\label{tab:related}
\centering
\begin{tabular}{l l l l }
	\toprule
	Model &  Scoring Function \\
	\midrule
	TransE~\cite{bordes2013translating} & $-\| \ve_0^h  - \ve_0^t + \vr_0 \|$ \\
	InterHT~\cite{wang2022interht} & $-\| \ve_0^h \circ \ve_1^t - \ve_1^h \circ \ve_0^t + \vr_0 \|$ \\
	TripleRE~\cite{long2021triplere} & $-\| \ve_0^h \circ \vr_1 - \ve_0^t \circ \vr_2 + \vr_0 \|$ \\
	PairRE~\cite{chao2020pairre} & $-\| \ve_0^h \circ \vr_1 - \ve_0^t \circ \vr_2 \|$ \\
	TranS~\cite{zhang2022trans} & $-\| \ve_0^h \circ \ve_1^t - \ve_1^h \circ \ve_0^t + \vr_0 + \ve_0^h \circ \vr_1 + \ve_0^t \circ \vr_2 \|$ \\
	\midrule \midrule
	AutoWeird & $-\| -\ve_1^t \circ \vr_2 + \ve_0^t \circ \vr_0 + \ve_0^t \circ \vr_2 - \vr_0\|$ \\
	\bottomrule 
\end{tabular}
\end{table*}

This search space covers a large number of different scoring functions. 
Note that with 7 different vectors $\ve_0^h, \ve_1^h, \vr_0, \vr_1, \vr_2, \ve_0^t, \ve_1^t$, 
the total number of first-order and second-order terms is $7+7 \times 7=56$. 
And for each of these 56 terms, we have three possible choices for its coefficient: 1, -1 and 0, 
where 0 coefficient means that it does not appear in $f(\cdot)$. 
Therefore, the total number of possible scoring functions can be as large as $3^{56} \approx 5.23 \times 10^{26}$.

\subsection{Search algorithm}
\label{ssec:alg}

To find new scoring functions from this search space, we consider the following bi-level optimization problem: 
\begin{align*}
\min_{\bm{\alpha}} \mathcal{M}_{\text{val}}
\left( 
\{ \ve(\bm{\alpha}), \vr(\bm{\alpha}) \} 
\right)
\quad \text{s.t.} \quad 
\{ \ve(\bm{\alpha}), \vr(\bm{\alpha}) \} = \min_{\{ \ve, \vr \}} \mathcal{L}_{\text{tr}}(\{ \ve, \vr \}; s_{\bm{\alpha}}),
\end{align*}
where $\bm{\alpha}$ denotes the hyper-parameters that specify the SF $s_{\bm{\alpha}}$, 
$\mathcal{M}_{\text{val}}$ refers to the validation performance (e.g., MRR on the validation set), 
and $\mathcal{L}_{\text{tr}}$ refers to the training loss. 
We use the self-adversarial negative sampling loss~\cite{sun2019rotate}, 
which is defined as follows:  
\begin{equation}
\begin{aligned}
\mathcal{L}(\{ \ve, \vr \}; s_{\bm{\alpha}}) 
= - \log\ \sigma(\gamma-s_{\bm{\alpha}}(h,r,t)) 
- \sum_{i=1}^{n} \frac{1}{k} \log\ \sigma (s_{\bm{\alpha}}(h'_i,r,t'_i)-\gamma), 
\end{aligned}
\end{equation}
where $\gamma$ is a fixed margin, $\sigma$ is the sigmoid function, and $(h'_i, r, t'_i)$ is the $i$-th negative triplet.

For simplicity, we use random search to solve this problem. 
We first restrict the $f(\cdot)$ function to be computed by 5 first-order or second-order terms, 
and randomly choose 4 terms among all 56 terms. 
Then for each term, we randomly assign a coefficient 1 or -1 for it.
The final function $f(\cdot)$ is computed by summing over these terms multiplied by their coefficients. 
We repeat the random search for 10 times. 
The searched SF, referred as AutoWeird, is also shown in Table~\ref{tab:related}.
Different with all previous scoring functions, it only uses information from the relation ($\bm r_0, \bm r_2$)
 and tail entity ($\bm e_0^t$  and $\bm e_1^t$) 
 to compute the score. 
Intuitively, it should have bad performances as it does not consider the head entity. 
However, as will be demonstrated in experimental parts, 
it can achieve state-of-the-art performances on some KG data sets. 

\section{Experiments}
\label{sec:exp}

In this section, we will first introduce the dataset.
Then we will introduce implementation details, 
experimental results,
and the post-analysis on the results.

\subsection{Dataset and metric}

The ogbl-wikikg2~\cite{hu2020ogb} is a large KG dataset extracted from Wikidata~\cite{vrandevcic2014wikidata}.
It contains a set of triplet edges, capturing the different types of relations between entities in the world. 
There are $2,500,604$ entities, $535$ relation types and $17,137,181$ triplets in the whole data set. 

We use original dataset splitting configuration. It splits the triplets according to time, simulating a realistic KG completion scenario that aims to fill in missing triplets that are not present at a certain timestamp. The training set contains 16,109,182 triplets, the validation set contains 429,456 triplets, and the test set contains 598,543 triplets.

Following the official guideline, we evaluate the KG embedding performance 
by predicting new triplet edges according to the training edges. 
The evaluation metric follows the standard filtered metric  \cite{bordes2013translating} 
which is widely used in KG.
Specifically, each test triplet edges are corrupted by replacing its head or tail with randomly-sampled 500 negative entities, 
while ensuring the resulting triplets do not appear in KG. 
The goal is to rank the true head (or tail) entities higher than the negative entities, 
which is measured by Mean Reciprocal Rank (MRR).

\subsection{Implementation details}

In our experiments, the Adam~\cite{kingma2014adam} optimizer is used with a learning rate of $0.0005$. 
The batch size of the model is set to 512. 
To prevent overfitting, we use dropout technique, and set it to 0.1. 
The negative sampling size  is set to 128. 
And the dimension $d$ of each embedding vector $\ve_0, \ve_1, \vr_0, \vr_1, \vr_2 \in \R^{d}$ is set to 200. 
The maximum number of training steps is 300,000. 
We validate the model every 20 thousand steps. 
The number of anchors for NodePiece are 20,000. 
And $\gamma$ in the loss function is set to 6. 
These hyper-parameters are selected according to the performance on the validation set.

\subsection{Results}

The results on ogbl-wikikg2 are shown in Table~\ref{tab:wikikg2}. 
By only using information from relation and tail entity, 
the AutoWeird model still achieves an MRR of 0.7368 on validation set and 0.7368 on test set, 
which outperforms all previous models on the ogbl-wikikg2 dataset. 
The revised version of AutoWeird model also has a performance better than all the previous methods. 
These counter-intuitive results demonstrate some hidden problems on
the evaluation of KG embedding models.
Hence, we conduct a post-analysis in the next subsection.

\begin{table*}[!tp]
	\caption{Results on the ogbl-wikikg2 dataset.}
	\label{tab:wikikg2}
	\centering
	\begin{tabular}{lcccc}
		\toprule
		Model & \#Params & \#Dims & Test MRR & Valid MRR \\
		\midrule
		TransE~\cite{bordes2013translating} & 1251M & 500 & 0.4256 $\pm$ 0.0030 & 0.4272 $\pm$ 0.0030 \\
		RotatE~\cite{sun2019rotate} & 1250M & 250 & 0.4332 $\pm$ 0.0025 & 0.4353 $\pm$ 0.0028 \\
		PairRE~\cite{chao2020pairre} & 500M & 200 & 0.5208 $\pm$ 0.0027 & 0.5423 $\pm$ 0.0020 \\
		AutoSF~\cite{zhang2020autosf} & 500M & - & 0.5458 $\pm$ 0.0052 & 0.5510 $\pm$ 0.0063 \\
		ComplEx~\cite{complex} & 1251M & 250 & 0.5027 $\pm$ 0.0027 & 0.3759 $\pm$ 0.0016 \\
		TripleRE~\cite{long2021triplere} & 501M & 200 & 0.5794 $\pm$ 0.0020 & 0.6045 $\pm$ 0.0024 \\
		\midrule
		ComplEx-RP~\cite{chen2021rp} & 250M & 50 & 0.6392 $\pm$ 0.0045 & 0.6561 $\pm$ 0.0070 \\
		AutoSF + NodePiece~\cite{galkin2021nodepiece} & 6.9M & - & 0.5703 $\pm$ 0.0035 & 0.5806 $\pm$ 0.0047 \\
		TripleREv2 + NodePiece~\cite{long2021triplere} & 7.3M & 200 & 0.6582 $\pm$ 0.0020 & 0.6616 $\pm$ 0.0018 \\
		InterHT + NodePiece~\cite{wang2022interht} & 19.2M & 200 & 0.6779 $\pm$ 0.0018 & 0.6893 $\pm$ 0.0015  \\
		TripleREv3 + NodePiece~\cite{long2021triplere} & 36.4M & 200 & 0.6866 $\pm$ 0.0014 & 0.6955 $\pm$ 0.0008 \\
		TranS + NodePiece~\cite{zhang2022trans} & 19.2M & 200 & 0.6882 $\pm$ 0.0019 & 0.6988 $\pm$ 0.0006 \\
		\midrule
		AutoWeird + NodePiece & 19.2M & 200 & \textbf{0.7353 $\pm$ 0.0006} & \textbf{0.7362 $\pm$ 0.0006} \\
		\midrule 
		EntOccur & - & - & 0.4419 & 0.4467 \\
		\bottomrule
	\end{tabular}
\end{table*}

\begin{table*}[!tp]
	\caption{Results on the ogbl-biokg dataset.}
	\label{tab:biokg}
	\centering
	\begin{tabular}{lcccc}
		\toprule
		Model & \#Params & \#Dims & Test MRR & Valid MRR \\
		\midrule
		TransE~\cite{bordes2013translating} & 188M & 500 & 0.7452 $\pm$ 0.0004 & 0.7456 $\pm$ 0.0003 \\
		RotatE~\cite{sun2019rotate} & 188M & 250 & 0.7989 $\pm$ 0.0004 & 0.7997 $\pm$ 0.0002 \\
		ComplEx~\cite{complex} & 188M & 250 & 0.8095 $\pm$ 0.0007 & 0.8105 $\pm$ 0.0001 \\
		PairRE~\cite{chao2020pairre} & 188M & 200 & 0.8164 $\pm$ 0.0005 & 0.8172 $\pm$ 0.0005 \\
		AutoSF~\cite{zhang2020autosf} & 93.8M & - & 0.8309 $\pm$ 0.0008 & 0.8317 $\pm$ 0.0007 \\
		TripleRE~\cite{long2021triplere} & 470M & 200 & 0.8348 $\pm$ 0.0007 & 0.8360 $\pm$ 0.0006 \\
		ComplEx-RP~\cite{chen2021rp} & 188M & 1000 & {0.8492 $\pm$ 0.0002} & {0.8497 $\pm$ 0.0002} \\
		AutoBLM+KGBench~\cite{zhang2022bilinear} & 192M & - & \textbf{0.8536 $\pm$ 0.0003}  & \textbf{0.8548 $\pm$ 0.0002} \\
		\midrule
		AutoWeird & 376M & 500 & 0.6755 $\pm$ 0.0004 & 0.6752 $\pm$ 0.0003 \\
		\midrule 
		EntOccur & - & - & 0.3387 & 0.3387 \\
		\bottomrule
	\end{tabular}
\end{table*}

\subsection{Post-analysis: why AutoWeird is effective?}
\label{sec:postans}

\begin{figure}[ht]
	\centering
	\subfigure[Training set of ogbl-wikikg2. \label{fig:all}]
	{\includegraphics[width=0.355\textwidth]{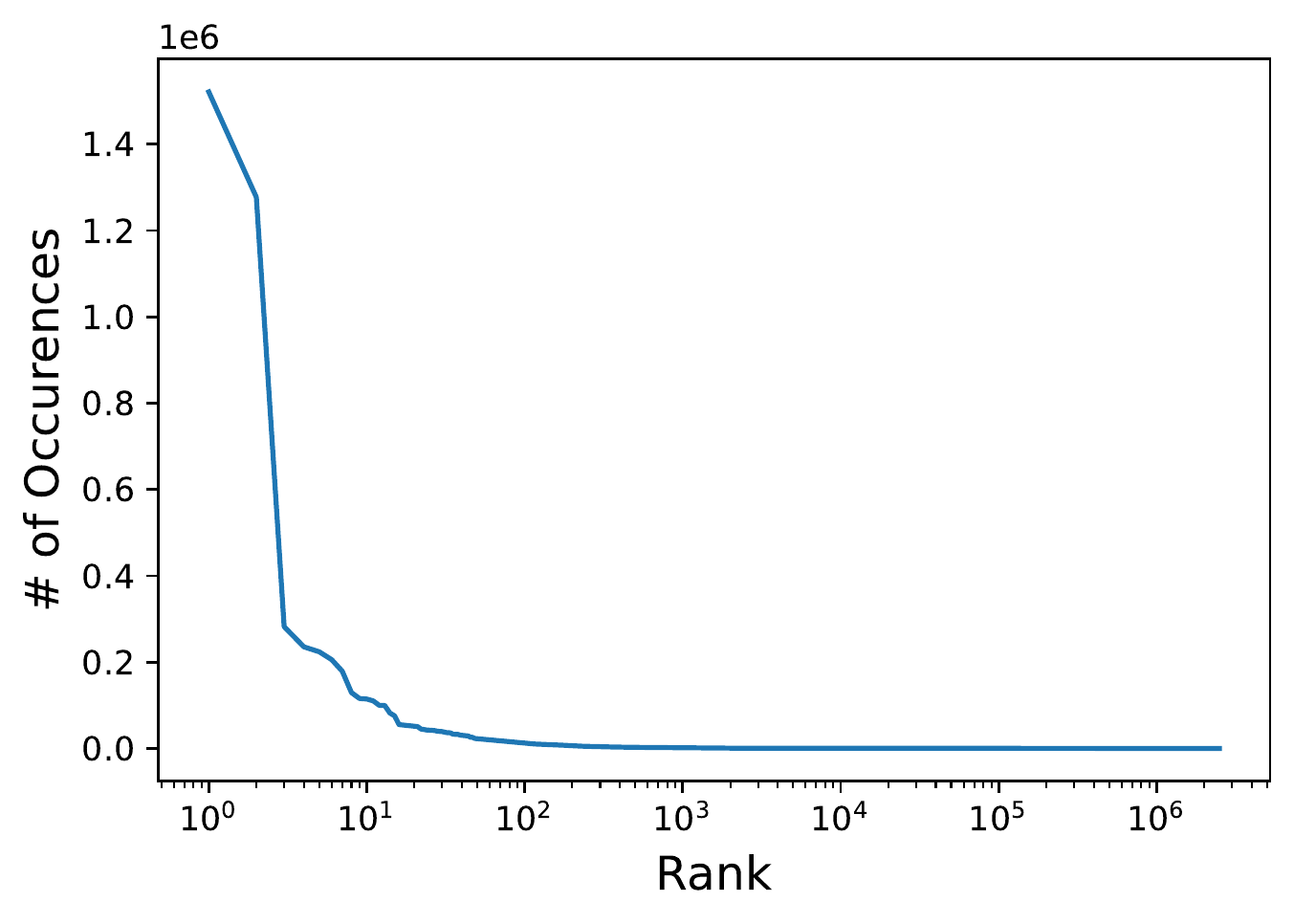}}
	\quad
	\subfigure[Testing set of ogbl-wikikg2. \label{fig:test}]
	{\includegraphics[width=0.37\textwidth]{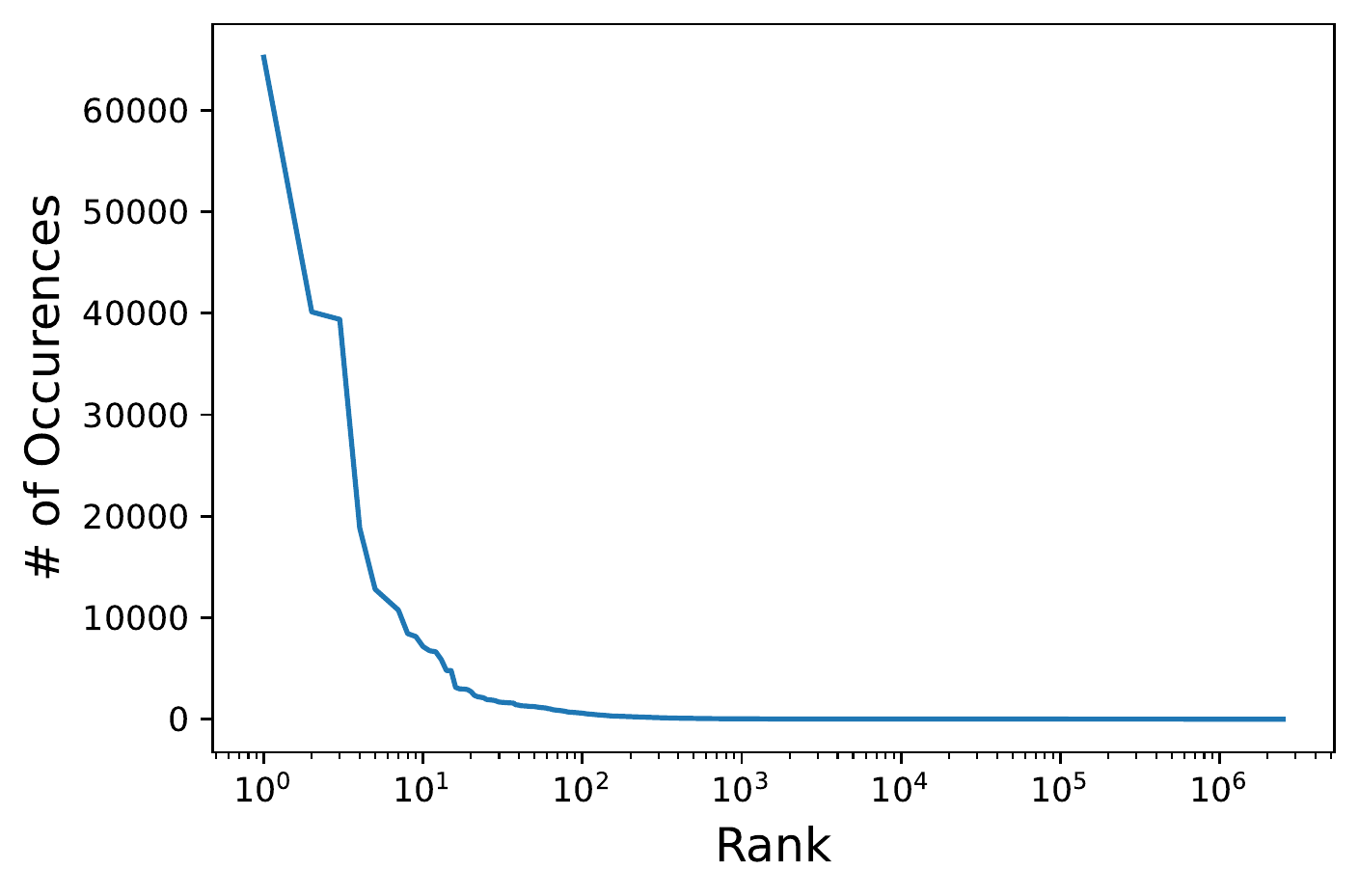}}
	\subfigure[Training set of ogbl-biokg. \label{fig:btr}]
	{\includegraphics[width=0.355\textwidth]{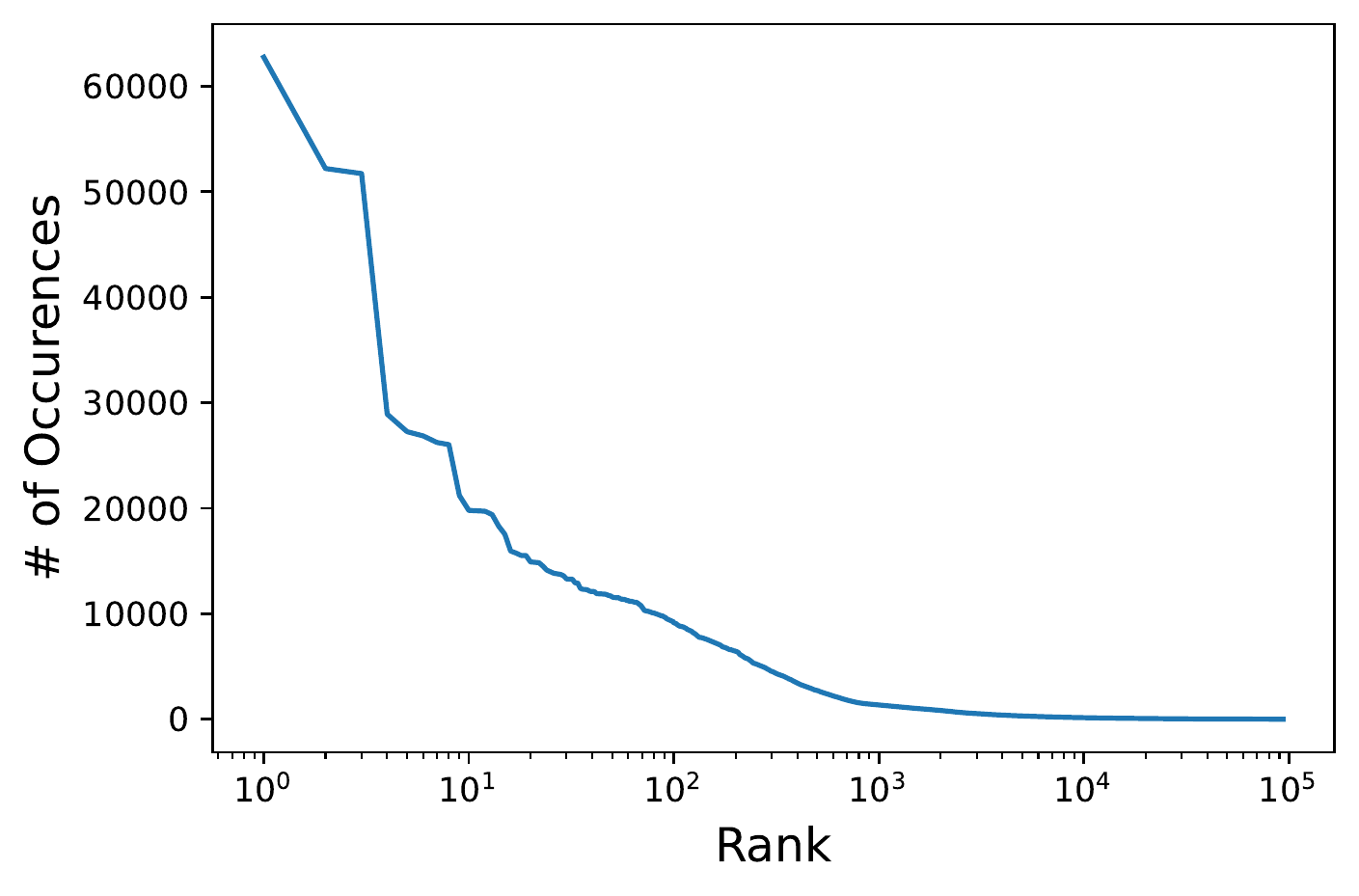}}
	\quad
	\subfigure[Testing set of ogbl-biokg. \label{fig:bte}]
	{\includegraphics[width=0.37\textwidth]{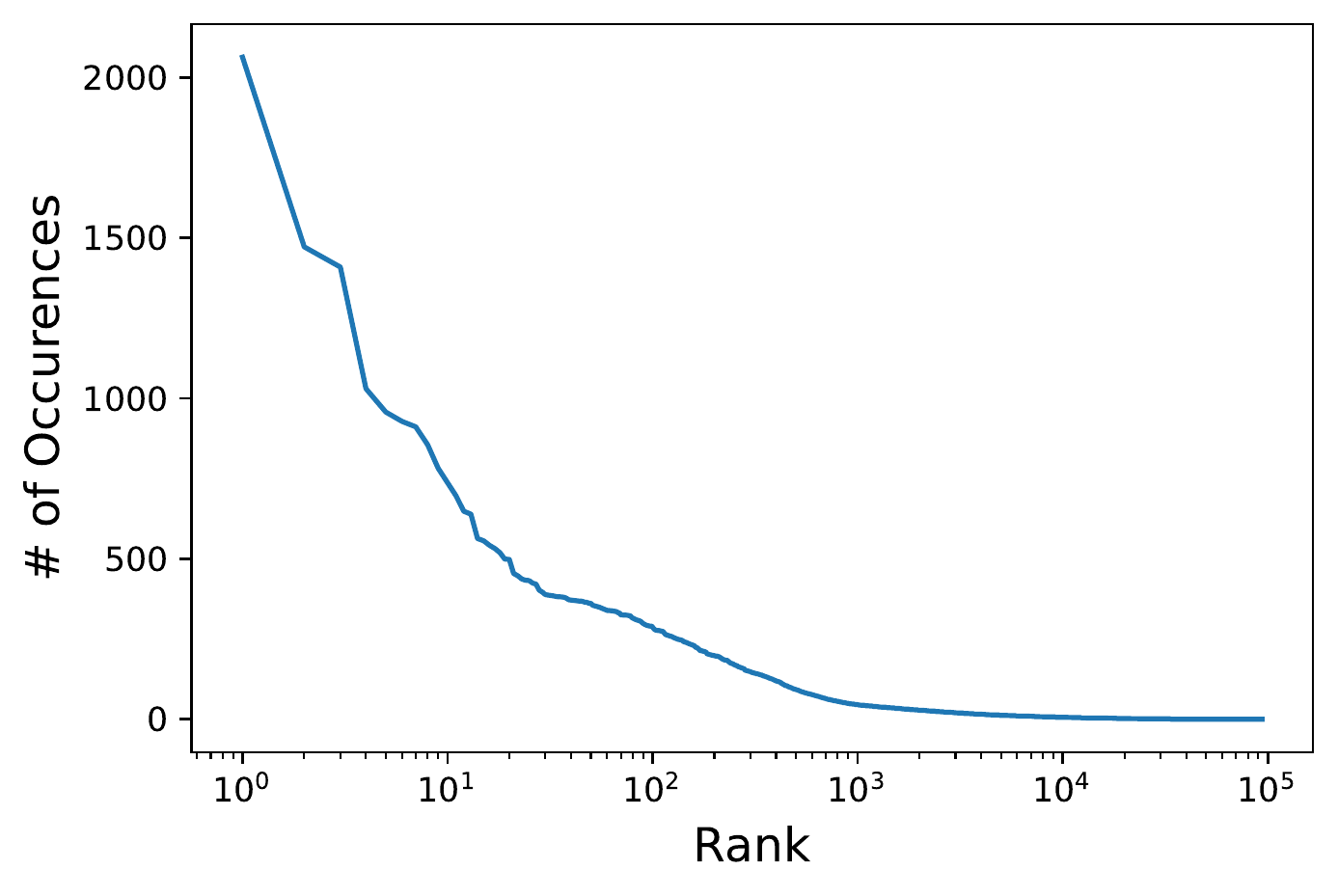}}
	\caption{Tail entity occurrence distribution for different KG data sets.}
	\label{fig:dist}
\end{figure}

To analyze why AutoWeird is effective,
we first count the occurrence of tail entities 
after adding inverse relations to the data set. 
Specifically,
for each entity,
we count how many times it appears as the 
tail entity
\footnote{ For simplicity, we regard the head prediction as tail prediction
	by adding inverse relations,
	i.e., $(t,r_\text{inv},h)$ for $(h,r,t)$ in the datasets.}
in the training set. 
As is shown in Figure~\ref{fig:all},
a few number of entities has more than one million occurrences in the training set.
Such a concentrated distribution also exists in the test triplets (Figure~\ref{fig:test}),
leading to a strong bias to the high occurrence entities.

Note that
the evaluation protocol in ogbl-wikikg2 is
to rank the positive tail entities
over a set of 500 randomly sampled negative entities.
Hence,
the bias will become more severe during evaluation.
As in Figure~\ref{fig:all} and \ref{fig:test},
if we only randomly sample $500$ entities from $10^6$ entities,
the majority of them will has low occurrence.

Based on these observations,
we design a simple method,
named as EntOccur,
by using the occurrence as scores.
For each triplet $(h,r,t)$ in the training set, 
we count the occurrence of entities $t$ for each relation $r$.
Then the score of $(h,r,t)$ in inference is directly 
set as the occurrence of $t$ with respect to 
the specific relation $r$.
As in Table~\ref{tab:wikikg2},
such a simple model EntOccur 
outperforms some embedding models 
like TransE and RotatE.

We further evaluate the searched SF AutoWeird 
and the simple solution EntOccur on the other KG link prediction dataset
ogbl-biokg \cite{hu2020ogb}.
By comparing the performance of AutoWeird and EntOccur with the results on the leaderboard
\footnote{\url{https://ogb.stanford.edu/docs/leader_linkprop/#ogbl-biokg}}
in Table~\ref{tab:biokg},
AutoWeird and EntOccur are much worse.
By comparaing ogbl-biokg with ogbl-wikikg2, we find that
\begin{itemize}
	\item the distributions of tail entity occurrence in ogbl-biokg in Figure~\ref{fig:btr} and \ref{fig:bte} are not as concentrated as those of ogbl-wikikg2 in Figure~\ref{fig:all} and \ref{fig:test};
	\item the number of entities in ogbl-biokg is $93,773$, much smaller than ogbl-wikikg2 with $2,500,604$;
	\item the negative samples for evaluation in ogbl-biokg 
		have the same type with the positive tail entity.
\end{itemize}
Hence, the negative entities in ogbl-biokg have higher correlation with the positive tail entity
than ogbl-wikikg2,
alleviating the bias towards entities with high occurrence.
As such, we attribute the success of AutoWeird in ogbl-wikikg2 
to the inappropriate evaluation protocol.

In conclusion, these results show that
there is a relatively strong correlation between relations and a few high occurrence entities in ogbl-wikikg2.
And the correlation may lead to the weird scoring function, AutoWeird in Table~\ref{tab:related}.
A better way to compare the different models
may be 
(1) using the full set of negative entities,
or (2) using some highly correlated negative entities,
or (3) using the larger number of negative entities
rather than $500$ randomly sampled ones.
However,
due to the time complexity in evaluating millions of negative entities for each triplet in ogbl-wikikg2, 
we leave it as a future work.

\section{Conclusion}
In this paper, we propose a novel search space for the scoring function in knowledge graph embedding model. 
We find a novel scoring function from this search space by random search, which only depends on tail entity and relation. 
Empirical results on the ogbl-wikikg2 dataset demonstrate that the searched scoring function is able to achieve state-of-the-art results among all previous models. 
These strange results should motivate us to consider potential problems 
on the evaluation of KG embedding models. 

\bibliography{anthology}

\begin{thebibliography}{10}

\bibitem{bordes2013translating}
A.~Bordes, N.~Usunier, A.~Garcia-Duran, J.~Weston, and O.~Yakhnenko.
\newblock Translating embeddings for modeling multi-relational data.
\newblock {\em Advances in neural information processing systems}, 26, 2013.

\bibitem{chao2020pairre}
L.~Chao, J.~He, T.~Wang, and W.~Chu.
\newblock Pairre: Knowledge graph embeddings via paired relation vectors.
\newblock {\em arXiv preprint arXiv:2011.03798}, 2020.

\bibitem{chen2021rp}
Y.~Chen, P.~Minervini, S.~Riedel, and P.~Stenetorp.
\newblock Relation prediction as an auxiliary training objective for improving
  multi-relational graph representations.
\newblock {\em arXiv preprint arXiv:2110.02834}, 2021.

\bibitem{galkin2021nodepiece}
M.~Galkin, J.~Wu, E.~Denis, and W.~L. Hamilton.
\newblock Nodepiece: Compositional and parameter-efficient representations of
  large knowledge graphs.
\newblock {\em arXiv preprint arXiv:2106.12144}, 2021.

\bibitem{hu2020ogb}
W.~Hu, M.~Fey, M.~Zitnik, Y.~Dong, H.~Ren, B.~Liu, M.~Catasta, and J.~Leskovec.
\newblock Open graph benchmark: Datasets for machine learning on graphs.
\newblock {\em arXiv preprint arXiv:2005.00687}, 2020.

\bibitem{hutter2019automated}
F.~Hutter, L.~Kotthoff, and J.~Vanschoren.
\newblock {\em Automated machine learning: methods, systems, challenges}.
\newblock Springer Nature, 2019.

\bibitem{ji2021survey}
S.~Ji, S.~Pan, E.~Cambria, P.~Marttinen, and S.~Y. Philip.
\newblock A survey on knowledge graphs: Representation, acquisition, and
  applications.
\newblock {\em IEEE Transactions on Neural Networks and Learning Systems},
  2021.

\bibitem{kingma2014adam}
D.~P. Kingma and J.~Ba.
\newblock Adam: A method for stochastic optimization.
\newblock {\em arXiv preprint arXiv:1412.6980}, 2014.

\bibitem{sun2019rotate}
Z.~Sun, Z.-H. Deng, J.-Y. Nie, and J.~Tang.
\newblock Rotate: Knowledge graph embedding by relational rotation in complex
  space.
\newblock {\em arXiv preprint arXiv:1902.10197}, 2019.

\bibitem{complex}
T.~Trouillon, C.~R. Dance, {{\'E}}ric Gaussier, J.~Welbl, S.~Riedel, and
  G.~Bouchard.
\newblock Knowledge graph completion via complex tensor factorization.
\newblock {\em Journal of Machine Learning Research}, 18(130):1--38, 2017.

\bibitem{vrandevcic2014wikidata}
D.~Vrande{\v{c}}i{\'c} and M.~Kr{\"o}tzsch.
\newblock Wikidata: a free collaborative knowledgebase.
\newblock {\em Communications of the ACM}, 57(10):78--85, 2014.

\bibitem{wang2022interht}
B.~Wang, Q.~Meng, Z.~Wang, D.~Wu, W.~Che, S.~Wang, Z.~Chen, and C.~Liu.
\newblock Interht: Knowledge graph embeddings by interaction between head and
  tail entities.
\newblock {\em arXiv preprint arXiv:2202.04897}, 2022.

\bibitem{wang2017knowledge}
Q.~Wang, Z.~Mao, B.~Wang, and L.~Guo.
\newblock Knowledge graph embedding: A survey of approaches and applications.
\newblock {\em TKDE}, 29(12):2724--2743, 2017.

\bibitem{yao2018automl}
Q.~Yao and M.~Wang.
\newblock Taking human out of learning applications: A survey on automated
  machine learning.
\newblock {\em arXiv preprint arXiv:1810.13306}, 2018.

\bibitem{long2021triplere}
L.~Yu, Z.~Luo, D.~Lin, H.~Liu, and Y.~Deng.
\newblock Triplere: Knowledge graph embeddings via triple relation vectors.
\newblock {\em arXiv preprint arXiv:2112.0095}, 2021.

\bibitem{zhang2022trans}
X.~Zhang, Q.~Yang, and D.~Xu.
\newblock Trans: Transition-based knowledge graph embedding with synthetic
  relation representation.
\newblock {\em arXiv preprint arXiv:2204.08401}, 2022.

\bibitem{zhang2020autosf}
Y.~Zhang, Q.~Yao, W.~Dai, and L.~Chen.
\newblock Autosf: Searching scoring functions for knowledge graph embedding.
\newblock In {\em 2020 IEEE 36th International Conference on Data Engineering
  (ICDE)}, pages 433--444. IEEE, 2020.

\bibitem{zhang2022bilinear}
Y.~Zhang, Q.~Yao, and J.~T. Kwok.
\newblock Bilinear scoring function search for knowledge graph learning.
\newblock {\em IEEE Transactions on Pattern Analysis and Machine Intelligence},
  2022.

\end{thebibliography}
\bibliographystyle{abbrv}

%%%%%%%%%%%%%%%%%%%%%%%%%%%%%%%%%%%%%%%%%%%%%%%%%%%%%%%%%%%%

%%%%%%%%%%%%%%%%%%%%%%%%%%%%%%%%%%%%%%%%%%%%%%%%%%%%%%%%%%%%

\end{document}